\pdfoutput=1
\documentclass[11pt]{article}

\usepackage[final]{acl}

\usepackage{times}
\usepackage{latexsym}
\usepackage{amsmath}
\usepackage{amssymb}
\usepackage{pifont}
\newcommand{\xmark}{\ding{55}}%
\usepackage{float}
\usepackage[T1]{fontenc}
\usepackage[utf8]{inputenc}

\usepackage{microtype}

\usepackage{inconsolata}

\usepackage{graphicx}

\usepackage{multirow}
\usepackage{colortbl}
\usepackage{booktabs}

\title{ACCEPT: Adaptive Codebook for Composite and Efficient Prompt Tuning}

\author{
    \textbf{Yu-Chen Lin\textsuperscript{1, \thanks{Equal Contribution.}}},
    \textbf{Wei-Hua Li\textsuperscript{1, $^*$}},
    \textbf{Jun-Cheng Chen\textsuperscript{2}},
    \textbf{Chu-Song Chen\textsuperscript{1}}
\\
     \textsuperscript{1}National Taiwan University,
     \textsuperscript{2}Academia Sinica
\\
 \small{
   \textbf{Correspondence:} \href{mailto:chusong@csie.ntu.edu.tw}{chusong@csie.ntu.edu.tw}}
 }

\begin{document}
\maketitle
\begin{abstract}
Prompt Tuning has been a popular Parameter-Efficient Fine-Tuning method attributed to its remarkable performance with few updated parameters on various large-scale pretrained Language Models (PLMs). Traditionally, each prompt has been considered indivisible and updated independently, {\color{black}leading the parameters increase proportionally as prompt length grows.}
To address this issue, we propose \textbf{A}daptive \textbf{C}odebook for \textbf{C}omposite and \textbf{E}fficient \textbf{P}rompt \textbf{T}uning (\textbf{ACCEPT}). In our method, we refer to the concept of product quantization (PQ), allowing all soft prompts to share a set of learnable codebook vectors in each subspace, with each prompt differentiated by a set of adaptive weights. We achieve the superior performance on 17 diverse natural language tasks including natural language understanding (NLU) and question answering (QA) tasks by tuning only $0.3\%$ of parameters of the PLMs. Our approach also excels in few-shot and large model settings, highlighting its significant potential. Our code is available on \href{https://github.com/AI-Application-and-Integration-Lab/Accept}{GitHub.}
\end{abstract}

\section{Introduction}
\label{sec:Intro}
With the blooming of large language models, Parameter Efficient Fine-Tuning becomes an effective solution to leverage the power of pretrained language models (LMs). Among various approaches, Prompt Tuning (PT) has been recognized for its simplicity and efficacy by adding tokens in front of the inputs. Though prompting pretrained LMs with specific or human-designed instructions makes model transferable to downstream tasks, additional effort is needed for elaborating the prompts as the output produced are often sensitive to them.
To address this issue, learning the prompts becomes a solution.
Prompt tuning~\cite{pt:21}, Prefix tuning~\cite{prefix:21} and P-tuning~\cite{ptune:21} replace explicit instructions with continuous prompt embeddings and provide flexibilities for the pretrained models to adapt themselves with superior performance. Following the concept, ATTEMPT~\cite{attempt:22}, MPT~\cite{mpt:23}, DePT~\cite{dept:24}, and TPT~\cite{tpt:23} demonstrate the capability of learnable PT in both single and multitask training scenarios.

However, previous studies often treat the prompts as independent units in learning. Though the learned prompts can be further clustered for noise filtering \cite{vip:22}, the parameters needed for training are not reduced since learning occurs before clustering. In this work, we introduce a method that represents the prompt based on a set of \textit{learnable codewords}. All prompts share a codebook with $\mathcal{N}$ codewords. Compared with updating the prompts independently and thus preventing word embeddings from sharing information with each other, {\color{black} our codebooks are sharable across all prompts in a downstream task, making codebooks' parameters %count 
size independent of the prompt length.}

In addition, our approach does not follow the common practice of regarding each prompt as inseparable.
When treating a prompt as an indivisible word embedding, we may overlook the possibility that, say, certain words may align with other words in the first half of the embedding and match different words in the second half. 
To tackle this issue, we adopt the idea of product quantization (PQ) \cite{jegou2010product} by dividing a prompt's word embedding into several subsections and construct a codebook for each subsection. In the past, PQ is effective for approximate nearest-neighbor search \cite{jegou2010product,yu2018product} and neural network compression \cite{wu2016quantized}. 
However, if we directly apply PQ to the learned parameters, their amount will not be lowered for training.
Hence, we simply follow PQ's concept where the codebooks are subsection-specific, and provide a set of learnable codewords for each subsection. This makes prompts share some identical subvectors, which allows part of tokens to have the same characteristics in a more fine-grained dimension.

To ease the learning process and make it differentiable, we allow each subvector to be softly combined (via linear coefficients) with the codewords, rather than being assigned by only one of the codewords as in PQ. This increases both diversity and flexibility of the representation. In the past, learning-based PT can be typically done by increasing the input length with prepended prompts %\cite{pt:21, attempt:22, tpt:23}
\cite{pt:21, attempt:22, tpt:23}
, or further adding the original embedding of words with the same number of additional prompts \cite{dept:24}. Our method, referred to as \textbf{ACCEPT}, is generally applicable and works for both.
We conduct experiments on 17 natural language tasks and show that our method consistently outperforms previous PT approaches.

\begin{figure*}[ht!]
\centering
\includegraphics[width=1.\linewidth]{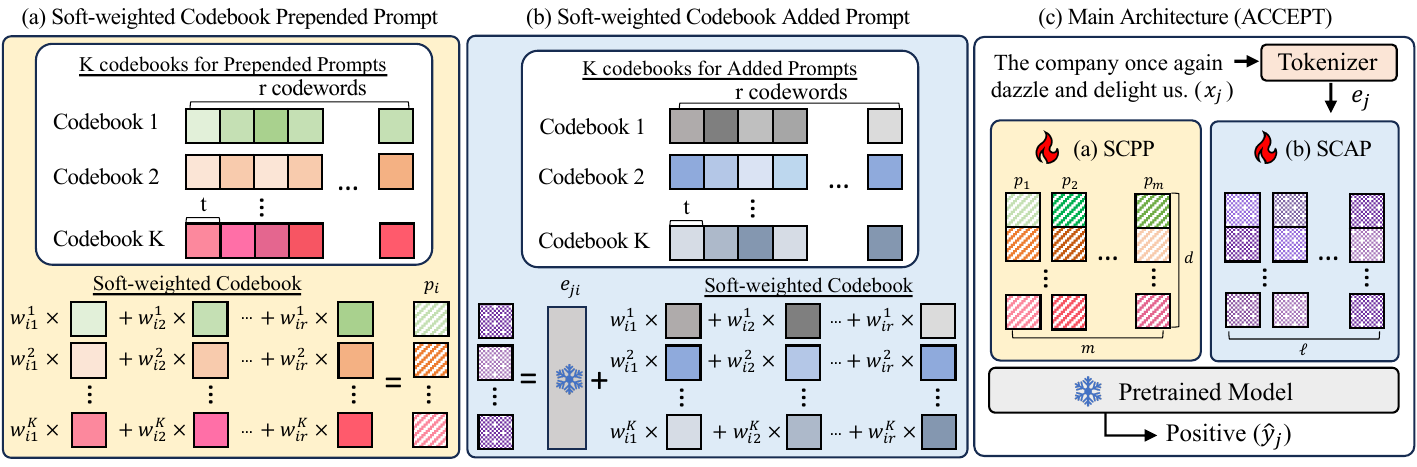}
  \caption{The overall model architecture of \textbf{ACCEPT}. 
  We subdivide both (a) \textit{S}oft-weighted \textit{C}odebook \textit{P}repended \textit{P}rompt (\textit{SCPP}) and (b) \textit{S}oft-weighted \textit{C}odebook \textit{A}dded \textit{P}rompt (\textit{SCAP}) to $K$ subspaces. Each subspace has a codebook with $r$ codewords shared by all prompts. Each sub-prompt is linearly combined by the codewords and weights. (c ) In the main architecture of ACCEPT, the final input is formed by prepending \textit{SCPP} to the word embedding updated with \textit{SCAP}. The pretrained model, with its parameters fixed, learns to output correct labels through tunable \textit{SCPP} and \textit{SCAP}.}
  \vspace{-10pt}
  \label{fig:model}
\end{figure*}

\section{Related Work}
% 由 PEFT 引入 PT
Parameter-efficient fine-tuning enhances the capabilities of pretrained LMs by updating a small set of parameters. The approach varies, such as training extra modules~\citep{adapter:19,lst:22} or modifying specific parts like biases or attention weights~\citep{bitfit:22,lora:21}. Among these, as Prompt Tuning (PT) is popular for its simplicity and effectiveness, we focus on PT.
% How Many Data Points is a Prompt Worth? => Prompt tuning is better under data scarcity
\subsection{Prompt Tuning Methods}
% PT is a promising parameter efficient fine-tuning methods to leverage the power of PLMs.
This track focuses on enhancing the quality and efficiency of prompts.~\citet{hard1:21} and ~\citet{hard2:20} incorporate manually crafted instructions into the input sequence to provide task-specific guidance helping steer the model's output. When the instructions are well-designed, models with frozen parameters exhibit excellent performance. However, additional effort is required for human adjustment %human-adjusting 
since the output is sensitive to the prompts. To address the issue,%\citet{autoprompt:20},
~\citet{entailment:21} and~\citet{lmbff:21} further generate hard prompt templates by model automatically. Nonetheless, optimizing discrete prompts is %still 
challenging. Thus, Prompt Tuning~\cite{pt:21}, Prefix tuning~\cite{prefix:21}, and P-tuning~\cite{ptune:21} turn prompts into continuous vectors,
known as soft prompt, which are prepended to %appended in front of 
the word embeddings. The learnable prompts are trained with the pretrained LMs frozen. By turning discrete prompts to a continuous space, the optimization can be achieved by a simple gradient descent. Recently, \citet{trans:21} and SPoT~\cite{spot:21} explore the advantages of initializing prompts by pretrained ones from other tasks. They demonstrate that learning prompts on one or more source tasks, and subsequently utilizing these learned prompts as initializations for a target task, is notably effective. ATTEMPT ~\cite{attempt:22}, MPT~\cite{mpt:23} and TPT~\cite{tpt:23} further design various architectures for multitask transfer learning. On the other hand, DePT~\cite{dept:24} focuses on reducing the training and inference time by decomposing prompt as a shorter one and a low-rank matrix added on word embeddings. Nevertheless, earlier approaches treat prompt as monolithic units, causing the number of trainable parameters to increase linearly with to the prompt length. In contrast, our method introduces a shared codebook in each subspace, which remains unaffected by the prompt length and facilitates information sharing among different prompts.

\subsection{Quantization in NLP}
Vector quantization (VQ) is a related technique to PQ which is widely employed in NLP.
% Quantization has also been used in multiple natural language tasks. For extractive opinion summarization, ~\citet{extract:21} encode input sentences into multi-head representations and then map them to a mixture of the nearest discrete latent codes with quantizer. 
VQ provides an effective discretization of latent sentence representations, making it especially suitable for NLP tasks due to the inherently discrete nature of text, as demonstrated in \citet{vqvae:17}, \citet{theory:18}, \citet{unsupervised:19}, \citet{disentangle:21} and \citet{extract:21}.
VQ is also used in PT.~\citet{vip:22} initially train a contextualized prompt for each input and cluster them using VQ to reduce variance. 

However, in the previous approaches, the number of parameters remains substantial since the training of original representations occurs before clustering. Different from these methods, we introduce learnable codebooks and adaptive weights which enable end-to-end training, thereby maintaining parameter efficiency throughout the process.

% \begin{figure}[t]
% \includegraphics[width=\columnwidth]{latex/figs/spq.pdf}
%   \caption{Soft Product-Quantized Soft Prompt Tuning. Soft Product Quantization introduces additional weights to each prompt in each subspace. The weighted-sum of each codewords according to the weights for each prompt compose the final prompt.}
%   \label{fig:spq}
% \end{figure}

\section{Methodology}
We first give a preliminary of PT for downstream tasks and PQ, and then present our method.
%\subsection{Preliminaries}
%\noindent\textbf{Prompt Tuning.}
\subsection{Prompt Tuning for Downstream Tasks}
Given a pretrained LM with parameters $\theta$, we want to transfer it to a target task with the training data %$D = \{(x_j, y_j) | j = 1,2,...N\}$. 
$D = \{(x_j, y_j)\}|_{j=1}^{|D|}$.
We first map $x_j$ to {\color{black}a sequence of word embeddings} $e_j$ as input, where $e_j \in R^{l \times d}$, $l$ is the maximal input sequence length and $d$ is the %model dimension. 
word-embedding dimension.
PT %adds appends
prepends a set of trainable continuous prompt embeddings $P = \{p_1,p_2,...p_m\}$ ($p_i \in R^d$) %in front of 
to the input embeddings while keeping the pretrained model parameters $\theta$ fixed. The training goal is to maximize the output probability of target $y_j$ as below,
%\begin{equation}
%\label{eq:pt}
%\max_{P_{pre}}\sum^{|D|}_{j=1} p_{\theta}(y_j|[P_{pre}, e_j];\theta).
%\end{equation}
%where $n$ is the total number of training samples.
\begin{equation}
\label{eq:pt}
\max_{P}\sum^{|D|}_{j=1} p_{\theta}(y_j|\theta;[P,e_j]).
\end{equation}

%\noindent\textbf{Product Quantization.}
\subsection{Review of PQ and Method Motivation}
% vector quantization 是 product 的一種 special case

%To better understand product quantization, first we introduce the vector quantization, which is a special case of product quantization.
VQ is known as the process of mapping a vector $x$ to the closest codeword $c^*$ in a codebook %$C$. The mapped codeword $c^* \in C = \{c_1, c_2,...,c_R\}$, where $R$ is the number of codewords. 
$C = \{c_1, c_2,...,c_{N}\}$ containing $N$ codewords.
%The goal is to minimize the difference between original $x$ and the mapped codeword $c^*$, % as below,
%\begin{equation}
%\label{eq:vq}
  % \begin{aligned}
%  L = \sum_{n}||x - c^*||^2, c^* \in C,
%  % \end{aligned}
%\end{equation}
%where $||\cdot||$ is the L2 distance metric and $n$ is the total number of samples. 
As an extension, PQ divides the vector $x\in R^d$ %with dimension $d$ 
into $K$ subspace, $x=[x^1, x^2,...,x^K]$, with $d=tK$ and $x^k\in R^t$. 
Each subspace possesses a codebook $C^k$ which contains $N_k$ codewords of dimension $t$ for $k=1\cdots K$. 
PQ thus exploits the Cartesian product of the codeword sets,
\begin{equation}
C=C^1 \times C^2 \times \cdots \times C^K,
\end{equation}
to encode the vector $x$. 
The total number of codewords becomes $\mathcal{N}=\Pi_{k=1}^K N_k$ for entire space. 
When $K=1$, PQ degenerates to VQ.

PQ has the advantage of enabling more codewords for the representation of $x$ by consuming fewer parameters.
Eg., if $N_k$ is the same for all $k$, PQ can take the storage cost of only $O(tKN_k) = O(dN_k)$ to provide the codewords amount of $(N_k)^K$.
For VQ, however, only $N_k$ codewords are provided under the same storage cost, or the storage should be increased to $O(d(N_k)^K)$ to get the same amount of codewords.
Hence, PQ is more parameter-efficient and suitable for PT.
The codewords distributed in subsections can enrich the diversity and flexibility of representation for the sub-problem solving.

%is then considered as a concatenation of $[x^1, x^2,...,x^k]$, where $x^j \in R^t$ and $t = d / k$. Each subspace possesses a codebook $C_k$. The mapped codeword $c^*$ then becomes a Cartesian Product of k codewords from every codebooks as follows,
%\begin{equation}
%\label{eq:pq}
%  \begin{split}
%  L &= \sum_{n}||x - c^*||^2, c^* \in C, \\
%  & \text{s.t.} \ c^* \in c^1 \times ... \times c^k.  \\
%  \end{split}
%\end{equation}

%It is intuitive to partition this problem into k sub-problems, where each sub-problem is degenerate to the vector quantization problem with $x$ and $c$ of dimension $t$. 
%The total number of variations of $c*$ then becomes $R^k$, while requiring only $O(Rd/k \cdot k) = O(Rd)$ storage (for parameters).

However, as mentioned above, parameter-efficient learning is not attainable if we perform PT first and then PQ.
Hence, our method does not really do the `quantization' step but only takes PQ's idea of efficient representation and makes the codewords of all subspaces learnable for PT. 
Moreover, for each subspace, we do not use only one codeword to express the input $x$ for that subsection,
but softly combining the codewords with linear coefficients for a more precise representation.
Details are given below.

%%%%%%%%%%%%%%%%%%%%%%%%%%%%%%%%%%%%%%%%%%%
%%%%%%%%%%%%%%%%%%%%%%%%%%%%%%%%%%%%%%%%%%%

\subsection{Proposed Method -- ACCEPT}
%\noindent\textbf{Motivation.}
%The previous methods viewed each prompt as a single word embedding and as a whole, which is independent to other prompts and indivisible. Nevertheless, we supposed that tokens share the same characteristics in a more fine-grained dimension. Thus, we introduced Soft Product Quantization mechanism into prompt tuning, and proposed ACCEPT to learning prompts in a more flexible and efficient way. 

Previous methods often view each prompt as a single and indivisible word embedding, independent to other prompts. 
We suppose that tokens can share the same characteristics in a more fine-grained dimension. 
Our method leverages the concept of PQ and partition embedding space into $K$ smaller subspaces. 
The $k$-th subspace has a codebook $\mathcal{C}^k = \{c_1^k, c_2^k,...,c_r^k\}$ containing $r$ codewords of dimension $t$, with $t = d / K$. % (shown in Fig.~\ref{fig:model}). 
Specifically, the total $K$ codebooks are shared across all prompts. % within each subspace. 
%Each prompt select one codeword in each codebook and combines them into a complete prompt with dimension $d$. According to different codeword choices of prompts, there are up to $R^K$ distinct combinations.

%\noindent\textbf{Soft product-quantization in prompt tuning.}
%\label{spq}
%We leverage the concept of product quantization and partition embedding spaces into $K$ smaller subspaces. The k-th subspace has a codebook $\mathcal{C}_k = \{c_1^k, c_2^k,...c_r^k\}$, which consists of $r$ codewords with dimension $t$, where $t = d / K$ (shown in Fig. ~\ref{fig:model}). The total $K$ codebooks are shared across all prompts within each subspace. Each prompt select one codeword in each codebook and combines them into a complete prompt with dimension $d$. According to different codeword choices of prompts, there are up to $R^K$ distinct combinations.

%Though product quantization reduce parameters by sharing discrete codewords, this may limit the variation between prompts to some extend. Thus we further introduced the Soft Weight mechanism. 

Remember that there is a set of trainable prompts $P = \{p_1, p_2,..., p_m\}$ ($p_i \in R^d$) for a downstream task in PT.
Similarly, we divide each $p_i$ into $K$ sub-prompts $p_i = \{p_i^1, p_i^2,..., p_i^K\}$ ($p_i \in R^t$).
We assign a group of weights, $W_i = \{w_i^1, w_i^2,..., w_i^K\}$ ($w_i^k \in R^r$), to the $i$-th prompt in every subspace. 
% Each prompt varies by a specific weight $w$ for each codewords $c$ in every subspaces. 
%Rather than select a specific codeword, we multiply the weights to each codewords. The weighted sum then form the final sub-prompt in that subspace.  
%Specifically, for each downstream task, there is a set of trainable prompts $\textbf{P} = \{p_1, p_2,..., p_m, \forall p_i \in R^d\}$. 
%For each sub-prompt, %$p_i^k$ has a set of weight $w_i^k = \{w_{i1}^k, w_{i2}^k,..., w_{ir}^k\}$ for combining $r$ codewords. 
A sub-prompt $p_i^k$ is then expressed as a linear combination of the codewords in $\mathcal{C}^k$ using the coefficient weights $w_i^k = \{w_{i1}^k, w_{i2}^k,..., w_{ir}^k\}$. % for combining $r$ codewords. 
Thus, the $k$-th subvector of the $i$-th prompt is calculated as
\begin{equation}
\label{eq:spq}
  \begin{aligned}
 p_{i}^k &= c_1^k \times w_{i1}^k + c_2^1 \times w_{i2}^k + ... + c_r^k \times w_{ir}^k, \\
  \end{aligned}
\end{equation}
where $c^k_j$ is the $j$-th codeword ($j=1,\cdots,r$) in the $k$-th codebook and $w^k_{ij}$ is the weight for $c^k_j$ in the $i$-th prompt, respectively.
%The shared codewords served as bases learning common information between sub-prompts, while the weights allow them to retain unique characteristics instead of entirely confined to the co-learned codewords.
%{\color{red} Fig.~\ref{fig:model} shows how the prompts are calculated and combined.}
%Finally, $p_i$ is concatenated ($\oplus$) by $K$ sub-vectors as below,
%\begin{equation}
%\label{eq:prompt}
%p_i = p^1_i \oplus p^2_i %\oplus ... \oplus p^k_i, 
%\end{equation}
%Thus, the total $m$ prompts for each downstream task $P_{task} = [p_1, p_2,...,p_m]^T \in R^{m \times d}$. By sharing the same bases, prompts can exhibit common characteristics in some aspects, while maintaining diversity among each other.

The prompts in $P$ thus depend on both the sets of codewords, $\mathbf{C}=\{\mathcal{C}^k|k=1,\cdots,K\}$ and combination weights, $\textbf{W}=\{W_i|i=1,\cdots,m\}$, in our method. To reflect this, we denote it as $P(\mathbf{C},\mathbf{W})$ in the following.
Our goal is to maximize the output probabilities of the ground truth label $y_j$ as
\begin{equation}
\label{eq:prepend_prob}
\max_{\mathbf{C},\mathbf{W}}\sum^{|D|}_{j=1}p_{\theta}(y_j|\theta;[{P(\mathbf{C},\mathbf{W})}, {e_j}]), 
\end{equation}
where ${P(\mathbf{C},\mathbf{W})} \in R^{m \times d}$. We refer Eq.~\ref{eq:prepend_prob} to as the Soft-Weighted codebook Prepended Prompt (\textit{SCPP}) tuning %learning 
in our method, as the prompts are prepended %in front of 
to the inputs. Fig.~\ref{fig:model}(a) gives an illustration of \textit{SCPP} learning.

%\noindent\textbf{Added Prompt.}
% 需要提到 lora 就是一種 vector quantize 嗎
% Need to compare with DePT?
As mentioned in Sec.~\ref{sec:Intro}, PT can also be done by adding complementary prompts to the original embedding \citep{dept:24}, and our method works for both. 
To achieve this, we conduct another prompt set $Q$ that contains the same number of $(l)$ prompts of the same length ($d$) to the input word embedding $e_j$ ($e_j\in R^{l\times d}$).
Similar mechanisms are applied to $Q$, which depends on the learnable codewords $\mathbf{C'}$ and combination weights $\mathbf{W'}$ too. 
%We apply the same mechanism to the complementary prompts, 
%which depend on another sets of $\mathbf{C'}$ and $\mathbf{W'}$ and denoted as $Q(\mathbf{C'},\mathbf{W'})$. 
We optimize $\mathbf{C'}$ and $\mathbf{W'}$ by solving % maximizing the output probabilities with the updated word embedding as
\begin{equation}
\label{eq:added_prob}
\max_{\mathbf{C'},\mathbf{W'}}\sum^{|D|}_{j=1}p_{\theta}(y_j|\theta;[{e_j} +{Q(\mathbf{C'},\mathbf{W'})} ]),
\end{equation}
where ${Q(\mathbf{C'},\mathbf{W'})} \in R^{l \times d}$ having the same shape of $e_j$. 
We refer Eq.~\ref{eq:added_prob} to as the Soft-Weighted codebook Added Prompt (\textit{SCAP}) tuning%learning
, as the prompts are added to the original word embeddings as updates. Fig.~\ref{fig:model}(b) illustrates the \textit{SCAP} learning. 
%The \textit{SCPP} and word embedding updated with \textit{SCAP} 
Combing \textit{SCPP} and \textit{SCAP} then forms our final %input of
\textbf{ACCEPT} (shown in Fig.~\ref{fig:model}(c)). 
At the same scale of parameters, combining the two types of prompts reduces the total input length, which makes training and inference more efficient \cite{dept:24}. 
% under the same scale of parameters.
% While maintaining the same parameter level and $l$, using a smaller $m$ makes the total input length ($m+l$) smaller, making the training and inference more efficient. Without loss of generality, we refer to DePT~\citep{dept:24} and decompose soft prompt into a shorter PP and AP.
\textbf{ACCEPT} then learns %to maximize the output probabilities of ground truth label $y_j$ for downstream tasks as
by maximizing 
\begin{equation}
\label{eq:output_prob}
\small
\max_{\mathbf{C},\mathbf{W},\mathbf{C}',\mathbf{W}'}\sum^{|D|}_{j=1}p_{\theta}(y_j|\theta;P(\mathbf{C},\mathbf{W}), e_j+Q(\mathbf{C}',\mathbf{W}')] ).
\end{equation}
%
%\begin{equation}
%\label{eq:output_prob}
%\max_{P_{pre}, %P_{add}}\sum^{|D|}_{j=1}p_{\theta}%(y_j|\theta;[{P_{pre}}, {e_j}^{'}] ),
%\text{s.t.} \ {e_j}^{'} = e_j + {P_{add}}
%\end{equation}
%
% \begin{equation}
% \label{eq:output_prob}
% \max_{P_{pre}, P_{add}}\sum^{|D|}_{j=1}p_{\theta}(y_j|\theta;\langle {P_{pre}}, {e_j}^{'}\rangle ),
% \text{s.t.} \ {e_j}^{'} = e_j + {P_{add}}
% \end{equation}
With only $[\mathbf{C},\mathbf{W}]$, $[\mathbf{C'},\mathbf{W'}]$ trainable and $\theta$ frozen.
% , we transfer the PLMs to target tasks efficiently.

\noindent\textbf{Number of Parameters.}
\label{sec:num_param}
% 需要再強調我們共用 codebook 可省 param 嗎?
With the vanilla PT, for a model having embedding dimension $d$ and $m$ prompts, the number of parameters is $md$. 
As we subdivide the embedding into $K$ subspaces, each $t$-dimensional ($t = d / K$) subspace has $r$ codewords, and thus the number of parameters of each codebook is $rt$. Total $K$ codebooks then need $rtK$ parameters. 
As for the weights, each prompt has $r$ weights in $K$ subspaces, which %is total 
contains total $rK$ parameters. Multiplied by the number of prompts $m$ %, then form the
forms a total of $rmK$ parameters. Finally, the total parameters of our method is as below,
\begin{equation}
\underbrace{rtK}_{\#para. of codebook} + \underbrace{rmK}_{\#para. of weight} = rd + rmK.
\end{equation}
Note that the number of parameters for the codebook is independent of the number of prompts, preventing linear growth with $m$.
To maintain the same scale of the trainable parameters with vanilla PT for a better comparison, we set $r$ by %abiding the equation 
letting $rd + rmK \leq md$. This ensures the number of parameter usage is no greater than the vanilla PT. % method with the same $m$. 

%\section{Experimental Settings}
\section{Experiments}
We present the experimental results and comparisons to other approaches in this section.
%We present experimental details in Section~\ref{sec:exp_setting} and showcase our primary results in Section~\ref{sec:result}. We conduct few-shot experiments in Section~\ref{sec:exp_few_shot} and provide ablation studies in Section~\ref{sec:exp_abs}.

\subsection{Experimental Settings}\label{sec:exp_setting}

%\noindent\textbf{Datasets and Evaluation Tasks.}
\noindent\textbf{Datasets and Tasks.}
% 補: split 的方式
Following previous works, we evaluate our %proposed 
method on 13 NLU tasks and 4 QA tasks, including (1) MNLI \citep{mnli:17}, QQP, QNLI \citep{qnli:18}, SST-2 \citep{sst2:13}, STS-B \citep{stsb:17}, MRPC \citep{mrpc:05}, RTE \citep{rte:07} and CoLA \citep{cola:19} from GLUE \citep{glue:18} benchmark; (2) MultiRC \citep{multi:18}, BoolQ \citep{bool:19}, WiC \citep{wic:18}, WSC \citep{wsc:12} and CB \citep{cb:19} from SuperGLUE \citep{superglue:19} benchmark; %and 
(3) MRQA 2019 Shared Task \citep{mrqa:19}, including Natural Questions \citep{nq:19}, HotpotQA \citep{hp:18}, SearchQA \citep{sqa:17} and NewsQA \citep{news:16}. We use SciTail \citep{khot2018scitail} additionally for few-shot learning.

\noindent\textbf{Baselines.}
We compare the proposed approach with various PEFT baselines including: (1) Fully fine-tuning (FT), where all the parameters of the pretrained backbone models are updated; (2) Prompt Tuning (PT)~\cite{pt:21}, where prompts are initialized by randomly sampled top vocabularies; (3) Some variants of PT, changing prompt architectures or utilizing knowledge transfer from other tasks such as SPoT~\cite{spot:21}, ATTEMPT~\cite{attempt:22}, MPT~\cite{mpt:23}, TPT~\cite{tpt:23} and DePT~\cite{dept:24}; (4) Other PEFT methods including Adapter~\cite{adapter:19} and AdapterDrop~\cite{ruckle2021adapterdrop}, inserting lightweight modules in the middle blocks of the pretrained models; BitFit~\cite{bitfit:22}, updating the bias terms in the attention mechanism;  LoRA~\cite{lora:21}, updating the attention weights with two additional low-rank matrices; LST~\cite{lst:22}, transferring by a ladder-side network-based adapter; %architecture along-side; 
Hyperformer~\cite{hyperformer:21} and HyperDecoder~\cite{hyperdecoder:22}, training a module to output the weights of adapters. (IA)$^3$~\cite{liu2022few}, scaling activations by learned vectors.

\noindent\textbf{Models.}
To provide a fair comparison with the previous methods, the main experiments are performed on the T5-base~\cite{raffel2020exploring} model with 220M parameters and $d=768$. We also conduct experiments on other models with %different
various scales including T5-small, T5-large, T5-3B and Flan-T5-11B models with 60M, 770M, 3B, and 11B parameters, respectively. The model dimensions are $512$, $1024$, $1024$ and $1024$, respectively. Note that Flan-T5-11B is an enhanced version of T5 model that has been fine-tuned in a collection of tasks. 

\noindent\textbf{Implementation Details.}
In the main experiments on GLUE, SuperGLUE and MRQA datasets, we divide the embedding into $K=24$ and $K=2$ subsections for \textit{SCPP} and \textit{SCAP}, respectively, where the parameters are chosen based on the performances on a small dataset RTE (detailed in Sec.~\ref{abla:gran}). 
{\color{black}We primarily use a grid search to determine the learning rates (\textit{lr}) for both the codebook and weights in \textit{SCPP} and \textit{SCAP}. For \textit{SCPP}, the \textit{lr} searched are \{3e-1, 4e-1, 5e-1\}, while for \textit{SCAP}, we searched \{1e-4, 5e-4, 1e-3, 5e-3\}. Additionally, we observe that a larger lr is more suitable for \textit{SCAP} on the MRQA 2019 Shared Task. Therefore, we extend our search to include higher values $\{1, 5, 10\}$ for \textit{SCAP}. Note that for the experiments that train \textit{SCPP} or \textit{SCAP} alone, the backbone follows DePT (the length of prompt is $60$ and the rank of LoRA matrices equals to $30$). 

We train $30k$ steps for small datasets with less than $10k$ samples, and $300k$ steps for large datasets more than $10k$ samples, following~\citet{spot:21}.
We perform evaluations every 1,000 steps and save the best checkpoint based on performance on the evaluation dataset. The results on the test dataset are then reported using these best checkpoints. We choose a batch size of $16$ for T5-small, T5-base and T5-large models, $2$ for T5-3B and $1$ for Flan-T5-11B due to the GPU memory limitation. The warmup step and weight decay are $1,800$ and $0.01$, respectively. 
Experiments are conducted on a single Nvidia 3090 GPU with 24 GB memory or % multiple 
2 Nvidia V100 GPUs with 32 GB memory.}

We set the associated number of codewords $r$ to maintain the total number of parameters no bigger than PT with $m=100$. We conduct three initialization strategies for the codebooks and weights of \textit{SCPP} and \textit{SCAP} in ACCEPT: (1) randomly initialized, (2) initialized by the pretrained weights of the intermediate tasks (we use MNLI for GLUE/SuperGLUE task and SQuAD~\cite{squad:16} for QA tasks following~\citet{spot:21}), and (3) initialized by the target task itself. For the latter two strategies, we first train \textit{SCPP} and \textit{SCAP} and use the pretrained weights as initialization for ACCEPT. On each dataset, we select the best strategy as the final results following~\citet{tpt:23}. In the few-shot experiments, following ~\citet{hyperformer:21} and ~\citet{attempt:22}, we sample $\gamma = \{4, 16, 32\}$ training instances three times with different seeds and report the average and standard deviation of our results. \textit{SCPP} and \textit{SCAP} are pretrained with one of the selected source dataset (MNLI, QQP, SST-2, SQUAD~\cite{squad:16}, and ReCoRD~\cite{record:18}) following the previous methods ~\cite{trans:21},~\cite{attempt:22},~\cite{dept:24}. More detailed experiment setup is listed in Appendix~\ref{appendix:hyper}.

\definecolor{lightlightgray}{rgb}{0.85,0.85,0.85}
\newcolumntype{g}{>{\columncolor{lightlightgray}}c}

\begin{table*}
  \centering
  \tabcolsep=0.065cm
  \small
  \begin{tabular}{lgccccccccgcccccg}
    \hline
     &  & \multicolumn{9}{c}{\textbf{GLUE}} & \multicolumn{6}{c}{\textbf{SuperGLUE}} \\
    \cline{3-11} \cline{12-17}
    \multirow{-2}{*}{\textbf{Method}}&\multirow{-2}{*}{\textbf{\#Para.}}& MNLI & QQP & QNLI & SST-2 & STS-B & MRPC & RTE & CoLA & Avg. & Multi & Bool & WiC & WSC & CB & Avg. \\
    \hline 
    % \vspace{0.05cm}
    Fine-tuning   & 220M & \textbf{86.8} & \textbf{91.6} & 93.0 & 94.6 & 89.7 & 90.2 & 71.9 & 61.8 & 84.9 & 72.8 & 81.1 & 70.2 & 59.6 & 85.7 & 73.9     \\
    % \hdashline
    Adapter   & 1.9M & 86.5 & 90.2 & 93.2 & 93.8 & 90.7 & 85.3 & 71.9 & 64.0 & 84.5 & \textbf{75.9} & \textbf{82.5} & 67.1 & \textbf{67.3} & 85.7 & 75.7 \\
    AdapterDrop  & 1.1M & 86.3 & 90.2 & 93.2 & 93.6 & \textbf{91.4} & 86.3 & 71.2 & 62.7 & 84.4 & 72.9 & 82.3 & 68.3 & \textbf{67.3} & 85.7 & 75.3 \\
    BitFit    & 280K & 85.3 & 90.1 & 93.0 & 94.2 & 90.9 & 86.8 & 67.6 & 58.2 & 83.3 & 74.5 & 79.6 & 70.0 & 59.6 & 78.6 & 72.5 \\
    LoRA      & 3.8M & 86.3 & 89.0 & 93.2 & 94.3 & 90.9 & 90.1 & 75.5 & 63.3 & 85.3 & 72.6 & 81.3 & 68.3 & \textbf{67.3} & 92.9 & 76.5  \\
    LST    & 3.8M & 85.6 & 88.8 & 93.3 & 94.1 & 90.7 & 90.4 & 71.9 & 58.1 & 84.1 & – & – & – & – & – & –  \\
    HyperFormer(m) & 638K & 85.7 & 90.0 & 93.0 & 94.0 & 89.7 & 87.2 & 75.4 & 63.7 & 84.8 & 72.9 & \textbf{82.5} & 69.0 & \textbf{67.3} & 85.7 & 75.4 \\
    HyperDecoder(m) & 1.8M & 86.0 & 90.5 & \textbf{93.4} & 94.0 & 90.5 & 87.7 & 71.7 & 55.9 & 83.7 & 70.4 & 78.8 & 67.1 & 61.5 & 82.1 & 72.0 \\
    \hline
    PT     & 76.8K & 81.3 & 89.7 & 92.8 & 90.9 & 89.5 & 68.1 & 54.7 & 10.6 & 72.2 & 58.7 & 61.7 & 48.9 & 51.9 & 67.9 & 57.8  \\
    PT$^{\dag}$     & 76.8K & 83.4 & 90.2 & 93.1 & 91.9 & 90.2 & 90.1 & 78.8 & 60.7 & 84.8 & 65.7 & 63.7 & 50.8 & 51.9 & 67.9 & 60.0  \\
    SPoT & 76.8K & 85.4 & 90.1 & 93.0 & 93.4 & 90.0 & 79.7 & 69.8 & 57.1 & 82.3 & 74.0 & 77.2 & 67.0 & 50.0 & 46.4 & 62.9 \\
    ATTEMPT & 232K & 84.3 & 90.3 & 93.0 & 93.2 & 89.7 & 85.7 & 73.4 & 57.4 & 83.4 & 74.4 & 78.8 & 66.8 & 53.8 & 78.6 & 70.5 \\
    MPT & 77.6K & 85.9 & 90.3 & 93.1 & 93.8 & 90.4 & 89.1 & 79.4 & 62.4 & 85.6 & 74.8 & 79.6 & 69.0 & \textbf{67.3} & 79.8 & 74.1 \\
    DePT     & 76.8K & 85.0 & 90.4 & 93.2 & 94.2 & 90.8 & 90.7 & 79.1 & 63.8 & 85.9 & 74.3 & 79.3 & 68.7 & \textbf{67.3} & 92.9 & 76.5 \\
    TPT$^1$ & 539K & 85.5 & 90.1 & 93.2 & \textbf{94.7} & 89.8 & 89.7 & 82.3 & 59.8 & 85.6 & 74.4 & 80.1 & 69.8 & \textbf{67.3} & 94.6 & 77.2 \\
    \hline
    ACCEPT (Ours) & 74.9K & 85.9 & 90.4 & 93.3 & 94.5 & 91.0 & \textbf{93.1} & \textbf{86.3} & \textbf{68.8} & \textbf{87.9} & 74.9 & 82.3 & \textbf{70.5} & \textbf{67.3} & \textbf{96.4} & \textbf{78.3} \\
    \hline
  \end{tabular}
  
  \caption{Performance on GLUE and SuperGLUE with T5-base model. For comparisons with prior works, we use Pearson Correlation for STS-B, Matthews Correlation for CoLA, F1 for MultiRC (Multi), and accuracy for other tasks as metrics. $^1$ sourced from ~\citet{tpt:23} and the others sourced from~\citet{dept:24}. \dag The values are the improved results tuned by~\citet{dept:24}.$(m)$ refers to multi-task training.}
  \label{tab:glue&superglue}
\end{table*}

\begin{figure}[t]
\includegraphics[width=\columnwidth]{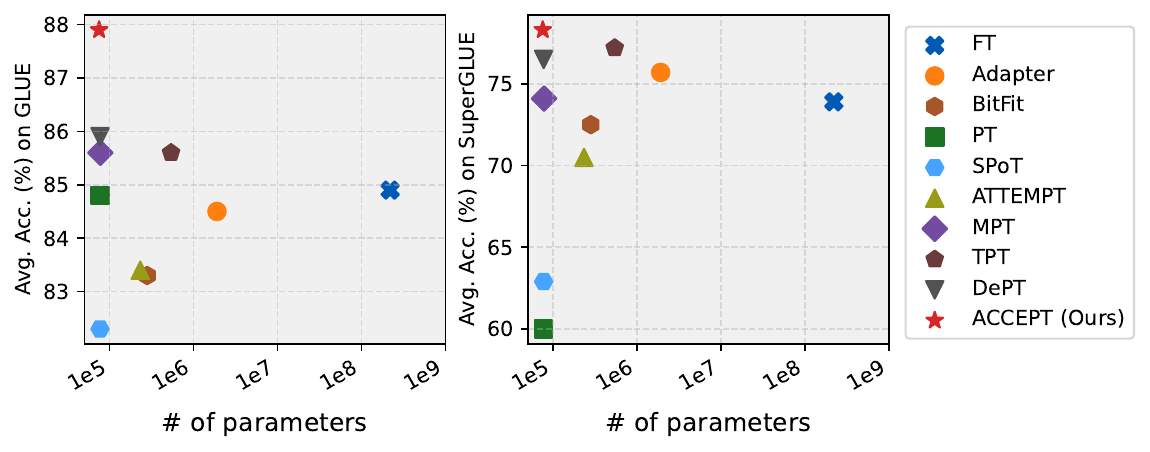}
  \caption{%The average 
  Average performance on the GLUE and SuperGLUE benchmarks relative to the number of trainable parameters for the T5-base model. ACCEPT achieves the best performance with the %least number of 
  fewest parameters.}
  \label{fig:performance_param}
\end{figure}

\begin{table}[t]
  \centering
  \tabcolsep=0.003cm
  \small
  {\resizebox{0.98\columnwidth}{!}{
  \begin{tabular}{lgccccg}
    \hline & & 
    \multicolumn{5}{c}{\textbf{MRQA}} \\
    \cline{3-7} \multirow{-2}{*}{\textbf{Method}}&\multirow{-2}{*}{\textbf{\#Para.}}
    & NQ & HP & SQA & News & Avg. \\
    \hline
    Fine-tuning   & 220M & 75.1 & 77.5 & 81.1 & 65.2 & 74.7 \\
    Adapter   & 1.9M & 74.2 & 77.6 & 81.4 & 65.6 & 74.7 \\
    BitFit  & 280K & 70.7 & 75.5 & 77.7 & 64.1 & 72.0 \\
    LoRA & 3.8M & 72.4 & 62.3 & 72.5 & 56.9 & 66.0 \\
    PT & 76.8K & 67.9 & 72.9 & 75.7 & 61.1 & 69.4 \\
    SPoT & 76.8K & 68.2 & 74.8 & 75.3 & 58.2 & 69.1 \\
    ATTEMPT & 232K & 70.4 & 75.2 & 77.3 & 62.8 & 71.4 \\
    MPT & 77.6K & 72.0$_{0.1}$ & 75.8$_{0.1}$ & 77.2$_{0.1}$ & 63.7$_{0.1}$ & 72.2 \\
    DEPT & 76.8K & 73.2$_{0.1}$ & 76.8$_{0.3}$ & 77.6$_{0.2}$ & 64.4$_{0.1}$ & 73.0 \\
    %
    % PQ-prompt & 74.5K & 73.8 & 76.7 & 79.1 & 64.5 & 73.5 \\
    % PQ-lora & 76.5K & 73.3 & 76.5 & 78.4 & 64.4 & 73.2 \\
    % ACCEPT & 74.2K & 73.6 & 77.0 & 78.9 & 64.6 & 73.5 \\
    % \textit{SCPP} & 74.5K & 73.6$_{0.3}$  & {76.9$_{0.01}$} & 78.7$_{0.4}$           & {64.7$_{0.1}$}           & 73.5 \\
    % \textit{SCAP}   & 76.5K & 73.4$_{0.2}$  & 76.8$_{0.5}$    & 78.5$_{0.2}$           & \color{red}64.3$_{0.07}$ & 73.3 \\     % Wei: PQ-lora (News) 訓練中
    \hline
    ACCEPT (Ours)      & 74.2K & 73.6$_{0.05}$ & 77.1$_{0.09}$   & {78.9}$_{0.01}$ & 64.6$_{0.06}$            & \textbf{73.6} \\
    \hline
    \end{tabular}
    }}
    \caption{Performance on the MRQA 2019 Shared Task. We report the average F1 score and standard deviation of three experiments with different seeds. The proposed method achieves promising performances with the limited number of parameters.}
  \label{tab:mrqa} 
\end{table}

%%%%%%%%%%%%%%%%%

\begin{table*}[h]
  \centering
  % \tabcolsep=0.1cm
  % \small
  {\resizebox{\linewidth}{!}{
  \begin{tabular}{l|ccccccccccc|c}
    \hline
    \multirow{2}{*}{\textbf{Task}} & $\gamma$-shot & FT & AD & PT & ST & HF & (IA)3 & ATP & MPT & TPT & DePT & ACCEPT (Ours) \\
    & \# Para. & 220M & 1.9M & 76.8K & 76.8K & 638K & 55.3K & 232K & 77.6K & 538K & 76.8K & 74.9K \\
    \hline 
    \multirow{3}{*}{BoolQ} & 4 & 50.5 & 53.4 & 61.6 & 50.5 & 48.0 & 56.7 & 61.8 & 62.2 & 62.2 & 62.7$_{5.4}$ & \cellcolor{lightlightgray}70.5$_{1.6}$\\
    & 16 & 56.5 & 51.4 & 61.9 & 50.6 & 50.2 & 62.0 & 60.0 & 63.3 & 63.5 & 66.9$_{4.4}$ & \cellcolor{lightlightgray}71.9$_{1.3}$\\
    & 32 & 58.4 & 54.5 & 61.7 & 61.2 & 58.3 & 67.2 & 65.3 & 68.9 & 67.4 & 67.2$_{3.4}$ & \cellcolor{lightlightgray}72.5$_{1.0}$\\
    \hline
    \multirow{3}{*}{CB} & 4 & 57.7 & 51.1 & 53.5 & 71.4 & 60.7 & 65.5 & \cellcolor{lightlightgray}82.1 & 73.6 & 78.6 & 75.0$_{5.1}$ & 78.6$_{3.6}$\\ 
    & 16 & 77.0 & 74.8 & 63.5 & 64.3 & 76.3 & 71.4 & 78.5 & 78.6 & 80.4 & 78.6$_{4.3}$ &\cellcolor{lightlightgray} 81.0$_{2.0}$\\
    & 32 & 80.0 & 74.8 & 67.8 & 64.3 & 81.4 & 75.0 & 85.7 & 82.1 & \cellcolor{lightlightgray}86.3 & 82.1$_{2.3}$ & 83.3$_{2.0}$\\
    \hline
    \multirow{3}{*}{SciTail} & 4 & 79.6 & 79.5 & 57.7 & 69.6 & 82.0 & 65.4 & 80.2 & 80.2 & \cellcolor{lightlightgray}81.0 & 78.1$_{2.5}$ & 79.0$_{4.4}$\\
    & 16 & 80.0 & 83.2 & 60.8 & 71.9 & 86.5 & 74.4 & 79.5 & \cellcolor{lightlightgray}87.3 & 85.5 & 78.5$_{1.4}$ & 80.5$_{3.1}$\\
    & 32 & 81.9 & 85.0 & 60.2 & 71.9 & 85.8 & 80.4 & 80.2 & \cellcolor{lightlightgray}86.3 & 85.2 & 85.4$_{3.1}$ & 84.8$_{0.4}$\\
    \hline
    \end{tabular}}}
    \caption{Few-shot learning results with $\gamma$ = \{4, 16, 32\} on BoolQ, CB, and SciTail datasets. FT: Fine-tuning, AD: Adapter, PT: Prompt tuning, ST: SPoT, HF: HyperFormer, ATP: ATTEMPT. ACCEPT significantly outperforms other methods on BoolQ and offers comparable performance on CB and SciTail with fewer parameters.}
  \label{tab:few_shot}
\end{table*}

\subsection{Results on NLU and QA Tasks}
% \subsection{Main Results}
\label{sec:result}
% {\color{red}Also MRQA results in this section.}
In Tab.~\ref{tab:glue&superglue}, we compare
the performances and the number of parameters during training of the proposed method with various methods on GLUE and SuperGLUE benchmarks. %with the number of parameters during their training. 
%According to the table, 
As can be seen, our method outperforms previous PT methods by a large margin,
% Compared to previous prompt tuning methods, our method outperforms with a large margin 
especially on MRPC, RTE and CoLA datasets of the GLUE benchmark, while consistently improving on other datasets such as MNLI, QQP, etc. 
Similar results can be found on the SuperGLUE benchmark. Our method achieves a great improvement on the Bool, WiC and CB datasets, while also yielding promising performances on MultiRC and WSC.
It is worth noting that %we 
our method %also 
surpasses previous PEFT methods
% designs 
exploiting much more %larger number of 
tunable parameters such as Adapter, %. Our method even 
and also outperforms FT %fully-finetuning 
by $3.0\%$ and $4.4\%$ on the average performances of GLUE and SuperGLUE with only $0.3\%$ parameters tuned.
We %also 
further visualize the average performances against the number of trainable parameters for each method in Fig.~\ref{fig:performance_param}. Our approach achieves the highest average accuracy while using the fewest parameters, making it more suitable for both performance and parameter efficiency than the others. %the most performance-parameter efficient method.

Besides having %outstanding 
favorable results on the NLU tasks above, the proposed method also %obtains great 
achieves nice performances on QA tasks. Tab.~\ref{tab:mrqa} demonstrates that our method achieves a $4.2\%$ improvement on the average of MRQA 2019 Shared Task than PT with fewer %less
parameters, further reducing %shortened 
the performance gap between FT and PT methods.

To conclude, the proposed method achieves state-of-the-art performances on the challenging GLUE/SuperGLUE benchmarks and MRQA 2019 Shared Task with fewer %less 
trainable parameters, highlighting %ours
its efficiency and effectiveness.

\subsection{Results on Few-shot Adaptation}\label{sec:exp_few_shot}
% \subsection{Few-shot Adaptations}\label{sec:exp_few_shot}
% ATTEMPT 寫說這個實驗是 follow hypernetwork
Following ~\citet{lmbff:21}, \citet{attempt:22}, ~\citet{mpt:23},
~\citet{tpt:23},
~\citet{dept:24}, we conduct the experiments with %the
a limited number of training samples available on the BoolQ, {\color{black}CB, and SciTail} datasets to verify the capability of ACCEPT in resource-limited scenarios. The experimental process involves initially training prompts on the intermediate tasks (e.g., MNLI) followed by transferring them to the target datasets with 4, 16, or 32 randomly sampled instances. %data.
In Tab.~\ref{tab:few_shot}, our method accomplishes impressive results %have excellent performances 
on BoolQ dataset, which is consistent with Tab.~\ref{tab:glue&superglue}. It also outperforms the previous methods on CB dataset with 4 shots. Note that for the CB dataset with 16 and 32 shots, our approach outperforms most of the methods except for ATTEMPT and TPT both using much more parameters than ours. The results demonstrate that ACCEPT remains effective %even 
in the few-shot adaptation scenarios.

%%%%%%%%%%%%%%%%%%%%%%%

\begin{table}[t]
  \centering
  \tabcolsep=0.05cm
  \small
  \begin{tabular}{cccc|c}
    \hline
    LC & PS & PP & AP & \textbf{SuperGLUE} \\
    \hline 
    % X & X & X & X & 57.8 \\
    \xmark & \xmark & \checkmark & \xmark & 60.0 \\
    \checkmark & \xmark & \checkmark & \xmark & 75.5 \\
    \checkmark & \checkmark & \checkmark &\xmark & \textbf{76.3} \\
    \hline
    \xmark & \xmark & \checkmark & \checkmark & 76.5 \\
    \checkmark & \xmark & \checkmark & \checkmark & 77.7 \\
    \checkmark & \checkmark & \checkmark & \checkmark & \textbf{78.3} \\ 
    \hline
    \end{tabular}
    \caption{Effectiveness of learnable codebook and subdivision. Our designs of shared learnable codebook (LC) % to a fine-grained level 
    and prompt embedding subdivision (PS) allow a performance gain with the same scale of parameters (76.8k) as other approaches. PP and AP denote the prepended and added prompt tunings, respectively.}
  \label{tab:abla_pq}
\end{table}

\subsection{Ablation Study}\label{sec:exp_abs}
\noindent\textbf{%Effectiveness of 
Learnable Codebook and Subdivision.}
To demonstrate the effectiveness of ACCEPT, we first conduct an ablation study of PQ, utilizing the shared learnable codebook and prompt embedding subdivision, with the prepended and added prompt tunings. Tab.~\ref{tab:abla_pq} shows that by sharing the learnable codebook among prompts, there is a noticeable performance improvement over the original architectures. Moreover, by dividing prompt embeddings into more fine-grained pieces, the performances are further enhanced. The results %demonstrate 
reveal the efficacy of PQ by subdividing the prompt embedding space.

% We first implement PT and DePT with non-divided codewords and weights. It is shown that by sharing the learnable codebook among prompts, the performance gain from the original architectures. Further, we conduct product quantization on PT and DePT to verify the power of divding prompt embeddings to a more fine-grained pieces. Tab.~\ref{tab:abla_pq} demonstrate that product quantization boost the performance from vector quantization, which prove the effectiveness of subdividing feature space.

%%%%%%%%%%%%%%%%%%%%%%%%

\begin{table}[t]
  \centering
  \tabcolsep=0.008cm
  \small
  \begin{tabular}{l|ccccgc}
    \toprule
    \multicolumn{7}{c}{\textbf{Soft-weighted Codebook Prepended Prompt (\textit{SCPP})}} \\
    \bottomrule
    \textbf{(t, r)} &
    % (8, 7) & (16, 12)  &
    \cellcolor{lightlightgray}(32, 20) & (64, 30) &
    % (96, 36) &
    (128, 40) &
    % (192, 45) &
    (256, 48) & (384, 51) & (768, 55) \\
    \hline 
    \textbf{\#Para.} & 
    % 76416 &  74496 & 
    \cellcolor{lightlightgray}74,880 & 75,360 &
    % 75648 &
    75,840 &
    % 76080 & 
    76,224 & 76,008 & 76,260 \\
    \hline
    \textbf{Acc.} & 
    % 77.70 & 77.70 &
    \cellcolor{lightlightgray}\textbf{82.73} & 79.14 &
    % \textbf{82.73} &
    79.14 & 
    % 81.29 &
    77.70 & \textbf{82.73} & 81.29 \\
    \toprule    
    \multicolumn{7}{c}{\textbf{Soft-weighted Codebook Added Prompt (\textit{SCAP})}} \\
    \bottomrule
    \textbf{(t, r)} & (32, 4) & (64, 8) &
    % (96, 10) &
    \cellcolor{lightlightgray}(128, 13) &
    % (192, 17) &
    \cellcolor{lightlightgray}(256, 20) & (384, 24) & (768, 30) \\
    \hline 
    \textbf{\#Para.} & 73,728 & 76,800 &
    % 74240 &
    \cellcolor{lightlightgray}76,032 &
    % 76544 &
    \cellcolor{lightlightgray}76,800 & 76,800 & 76,800 \\
    \hline
    \textbf{Acc.}
     & 77.70 & 76.98 &
     % 76.98 &
     \cellcolor{lightlightgray}79.86 &
     % 78.42 &
     \cellcolor{lightlightgray}81.29 & \textbf{82.73} & 78.42 \\
    \hline
    \end{tabular}
    \caption{Performance on RTE dataset with dividing \textit{SCPP} and \textit{SCAP} into different granularities. For T5-base, $t = 768$ means the prompt is NOT divided. Configs surpass non-division settings are highlighted in gray.}
  \label{tab:abla_gran}
\end{table}

\noindent\textbf{Different Granularity of Subdivision.}
\label{abla:gran}
% PQ better than VQ
% Find the best way to divide the word 
%Furthermore, we conduct the experiments to study 
We further study on the impact of using different sub-dimension ($t$) and codebook size ($r$) pairs in our approach. 
We choose multiple values for $K$ and divide the embeddings of dimension $d$ into multiple vectors with sub-dimension $t$. We then determine $r$ to satisfy $rtK + rmK \leq 100d$, ensuring fewer parameters are used compared to PT with $m=100$. Note that with T5-base model, $t = 768$ means no division on the embedding dimension. Tab.~\ref{tab:abla_gran} shows that with an appropriate division, there is a performance gain compared to treating the embedding as a whole. \textit{SCPP} achieves an optimal performance with $t=32$, chosen for its fewer parameters than $t=384$. For \textit{SCAP}, optimal performance is achieved with $t=384$. 
More complete experiments are in Appendix~\ref{appendix:abla_gran}.
%Optimal settings are applied to all datasets. 
Note that the optimal parameters ($t=32, K=24, r=20$ for \textit{SCPP} and $t=384, K=2, r=24$ for \textit{SCAP}) chosen from this small task (RTE) are then applied to ALL the datasets when using our approach with the T5-base model in the experiments.

\begin{figure}[t!]
\includegraphics[width=\columnwidth]{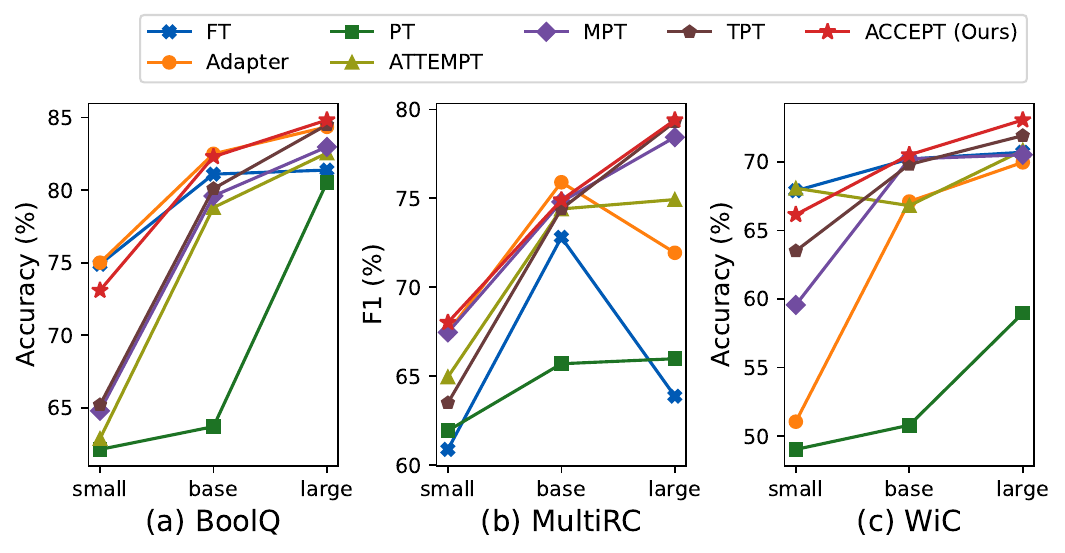}
  \caption{Performance on BoolQ, MultiRC and Wic datasets with different model sizes (T5-small, T5-base and T5-large). Our method shows improved performance as the model size increases and reaches SOTA on larger model, showcasing the potentional of ACCEPT.}
  \label{fig:model_size}
\end{figure}

\begin{table}
  \centering
  \tabcolsep=0.05cm
  \small
  \begin{tabular}{c c|c|c|c}
    \hline
    \textit{SCPP} &
    \textit{SCAP} &
    \textbf{GLUE} & \textbf{SuperGLUE} &
    \textbf{MRQA} \\
    \hline
    \xmark & \xmark & 85.9 & 76.5 & 73.0 \\
    \checkmark & \xmark & 87.1 & 77.8 & 73.5 \\
    \xmark & \checkmark & 87.0 & 77.5 & 73.3 \\ 
     \checkmark & \checkmark & \textbf{87.9} & \textbf{78.3} & \textbf{73.6} \\
    \hline
    \end{tabular}
    \caption{Effectiveness of Soft-weighted Codebook Prepended and Added prompts. The optimal performance is achieved by both \textit{SCPP} and \textit{SCAP} combined.}
  \label{tab:pre_add_abla}
  \end{table}

\noindent\textbf{Ablation on \textit{SCPP} and \textit{SCAP}.}
We train \textit{SCPP} and \textit{SCAP} individually, initializing them with a random Gaussian distribution. Tab.~\ref{tab:pre_add_abla} shows that when using \textit{SCPP} individually, the average performance on GLUE, SueprGLUE and MRQA improves $1.2\%$, $1.3\%$ and $0.5\%$, respectively. Similarly, %nature can be observed with \textit{SCAP}. There 
there are $1.1\%$, $1.0\%$ and $0.3\%$ performance gain %while introducing 
when using \textit{SCAP} individually. We find that the performance gain is relatively small on QA tasks. Improvement in the generation of longer sentences with ACCEPT is left as a future work. With both \textit{SCPP} and \textit{SCAP}, our approach achieves the best performances with $2.0\%$, $1.8\%$ and $0.6\%$ gain on each of the three benchmark, indicating the importance of both our designs. More detailed results on each dataset are in Appendix~\ref{appendix:abla_pq}.

%%%%%%%%
\begin{table}[t]
\begin{minipage}[t]{0.45\linewidth}
  \centering
  \tabcolsep=0.05cm
  \small
  \begin{tabular}{l|c}
    \hline
    \textbf{Method} & \textbf{GLUE} \\
    \hline 
    PT & 85.6 \\
    DePT & 86.4 \\
    TPT & 88.4 \\
    \hline
    ACCEPT (Ours) & \textbf{88.5} \\
    \hline
    \end{tabular}
    \caption{Performance on GLUE with T5-3B. We outperform all PT, DePT and achieve a 0.1 improvement over TPT with less parameters.}
  \label{tab:t5_3b}
\end{minipage}
\hspace{0.05\linewidth}
\begin{minipage}[t]{0.45\linewidth}
  \centering
  \tabcolsep=0.05cm
  \small
  \begin{tabular}{l|c}
    \hline
    \textbf{Method} & \textbf{RTE} 
    % & \textbf{MRPC} 
    \\
    \hline 
    PT$^{\dag}$ & 88.49 
    % & 92.16
    \\
    DePT$^{\dag}$ & 89.92 \\
    \hline
    ACCEPT (Ours) & \textbf{91.37} \\
    \hline
    \end{tabular}
    \caption{Performance of Flan-T5-11B on RTE dataset. Our method outperforms both PT and DePT. \dag The results are reproduced by us.}
  \label{tab:t5_11b}
\end{minipage}
\end{table}
%%%%%%%%%%%%%%%

\begin{table}[t]
\begin{minipage}[t]{1\linewidth}
  \centering
  \small
  \begin{tabular}{l|c|c}
    \hline
    \textbf{Method} & \textbf{\#Para.} & \textbf{SST-2} \\
    \hline 
    PT & 417.8K & 94.48 \\
    DePT & 413.4K & 94.95 \\
    \hline
    ACCEPT (Ours) & 405K & \textbf{95.64} \\
    \hline
    \end{tabular}
    \caption{Performance of Llama-2-7B model on SST-2 dataset. Our method outperforms PT and DePT. PT and DePT results are sourced from \cite{dept:24}.}
    
  \label{tab:llama2}
\end{minipage}
\end{table}

% \begin{table}[t]
% \begin{minipage}[t]{1\linewidth}
%   \centering
%   \small
%   \begin{tabular}{l|c|c}
%     \hline
%     \textbf{Method} & \textbf{\#Para.} & \textbf{Llama-3.1-8B}
%     % & \textbf{MRPC} 
%     \\
%     \hline 
%     PT$^{\dag}$ & 417.8K & 96.10
%     % & 92.16
%     \\
%     DePT$^{\dag}$ & 413.4K & 96.78 \\
%     \hline
%     ACCEPT (Ours) & 405K & {\color{red}\textbf{96.44}} \\
%     \hline
%     \end{tabular}
%     \caption{Performance of Llama-3.1-8B model on SST-2 dataset. Our method outperforms both PT and DePT. \dag The results are reproduced by us.}
%   \label{tab:llama3}
% \end{minipage}
% \end{table}
%%%%%%%%%%%%%%%

\noindent\textbf{Model Scaling.}
% 檢查圖片是否 consistent
We explore the effect of different model sizes (T5-small, T5-base and T5-large) with our method on BoolQ, MultiRC and WiC datasets in Fig.~\ref{fig:model_size}. Our method demonstrates increased performance improvement %effectiveness
with larger language model backbones, highlighting ACCEPT's adaptability with bigger models. We also provide the results of fully fine-tuning (FT), Adapter, Prompt Tuning (PT), MPT, and TPT for comparison. ACCEPT demonstrates competitive performance across all model scales. Notably, the tunable parameters of our approach are much fewer than those in FT, Adapter, and ATTEMPT. Despite this, we achieve state-of-the-art performance on all three datasets with T5-large (770M), which is a highly encouraging result given the reduced parameter count. 

To further study the capabilities and possibilities of ACCEPT on large language models, We conduct the experiments with billion-parameter %level 
models including T5-3B, Flan-T5-11B {\color{black}and Llama-2-7B.} % and Llama-3.1-8B}. 
Tab.~\ref{tab:t5_3b} shows that ACCEPT achieves the state-of-the-art average accuracy on GLUE benchmark with T5-3B. ACCEPT surpasses the vanilla PT and other prompt tuning methods including DePT and TPT. Notably, we achieve a $0.1\%$ improvement with much fewer parameters than TPT, which is an impressive result.
Flan-T5 is an enhanced version of T5 model by fine-tuning T5 on 1,800 downstream tasks. We further select the 11 billion-parameter version and investigate the effectiveness of ACCEPT on large language models. Due to the huge computation resource required, we select the RTE dataset for evaluation. Tab.~\ref{tab:t5_11b} shows that ACCEPT outperforms both PT and DePT on the RTE dataset. This indicates the potential and capability of ACCEPT incorporating large-scale models. 
{\color{black} We also evaluate our method using Llama-based models. Initially, we attempted to reproduce the results from ~\cite{dept:24} which uses the auto-regression generated output for classification. However, we found it challenging to achieve the same level of accuracy by this approach. To tackle this, we added a trainable linear head to output the probability distribution for classification. The results are shown in Tab. \ref{tab:llama2}. % and Tab. \ref{tab:llama3}. 
Our approach outperforms both PT and DePT on the SST-2 dataset by 1.16 and 0.69 with Llama-2-7B. %, and by {} and {} with Llama-3.1-8B, respectively. 
This demonstrates the excellent capability of our method with large language models (LLMs), highlighting its potential for adaptation 
to future LLM architectures.}
%%%%%%%%%%%%%%%%%%%%%%%

\begin{table}[t!]
  \centering
  \tabcolsep=0.08cm
  \small
  \begin{tabular}{l|c|c}
    \hline
    \textbf{Method} & \textbf{GLUE} & \textbf{SuperGLUE} \\
    \hline 
    PT$^\dag$ & 84.8 & 60.0 \\
    DePT & 85.9 & 76.5 \\
    \hline
    \textbf{Init. method of ACCEPT} \\
    \hline
    Random & \underline{87.1} & 76.5 \\
    Intermediate task & \textbf{87.5} & \textbf{77.6} \\
    Target task & \underline{87.1} & \underline{77.2} \\   
    \hline
    \end{tabular}
    \caption{Performance of ACCEPT on GLUE and SuperGLUE with different prompt initialization. All three strategies outperforms PT and DePT, showing our method's robustness.}
  \label{tab:init_glue_avg}
  \vspace{-10pt}
\end{table}

\noindent\textbf{Prompt Initialization.}
We further analyze how initialization affects the performance.
We conduct three initialization settings, (1) Random initialization: Both the codebooks and weights are initialized with a random Gaussian Distribution; (2) Intermediate task initialization: SPoT~\cite{spot:21} has shown that initializing prompts with the pretrained weights from the tasks of a similar nature can benefit the training of the target task. (3) Target task initialization: By first pretraining the codebooks and the weights of \textit{SCPP} and \textit{SCAP} respectively on the target task, both of them are then served as the initialization of ACCEPT. Tab.~\ref{tab:init_glue_avg} shows that our method achieves better performances than PT and DePT with all three strategies, revealing the robustness and effectiveness of ACCEPT. Moreover, intermediate task initialization strategy yields the best performances. We conjecture that the pretrained codebooks and weights from intermediate task of a similar nature helps the target task transfer more easily, providing additional knowledge and surpasses the performances of random or target task initialization. Detailed %performance 
results %for each dataset on the GLUE, SuperGLUE, and MRQA benchmarks 
are provided in Appendix~\ref{appendix:init}.

\begin{figure}[t!]
\includegraphics[width=\columnwidth]{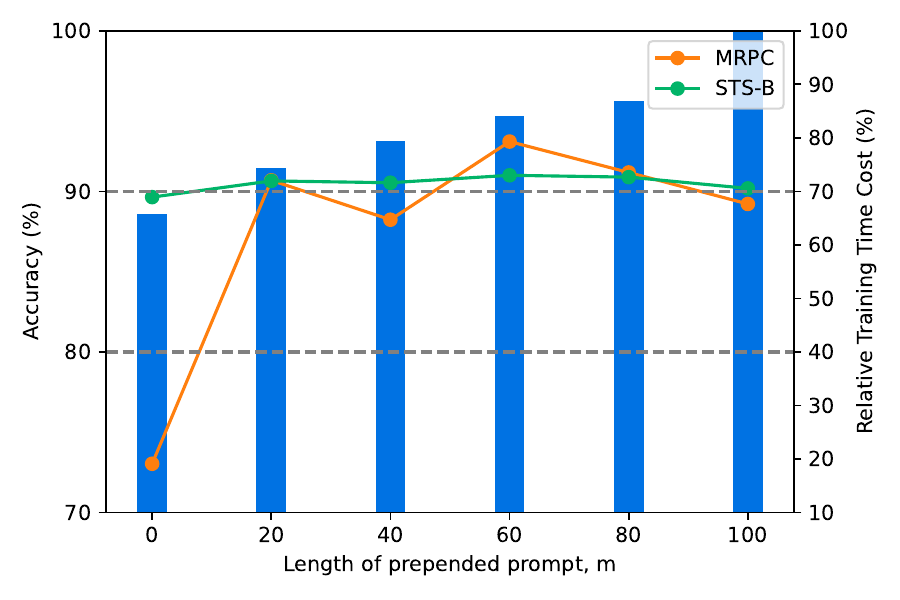}
  \caption{Performance on the MRPC and STS-B datasets and their relative training time (normalized to the one with $m = 100$) for various prompt lengths $m = \{20, 40, 60, 80, 100\}$. Both datasets show the best performance at $m = 60$.}
  \label{fig:prompt_length}
  \vspace{-10pt}
\end{figure}

\noindent\textbf{{\color{black} Prompt Length.}}%}
We evaluate the impact of different prompt lengths ($m$) on model performance and training time, as shown in Fig.~\ref{fig:prompt_length}. The experiments are conducted on the MRPC and STS-B datasets with $m$ values of $\{0, 20, 40, 60, 80, 100\}$, while maintaining the same level of training parameters across all settings. The results indicate that as $m$ increases, the training time also rises. Notably, our approach achieves peak accuracy in both datasets with $m = 60$, making it our optimal choice for the prompt length setting.

\section{Conclusion}
In this paper, we present ACCEPT, a novel prompt tuning method based on product quantization. As compared with other PT methods, the proposed method allows versatile and efficient prompt learning by subdividing prompt embeddings and computing each subprompt with the linear combination of learnable codewords and weights. 
Extensive experiments demonstrate that ACCEPT achieves outstanding performance across various NLP tasks. Furthermore, we also show the proposed approach is capable of being effectively adapted to billion-parameter models and achieves decent results.

% our method's adaptability to billion-parameter models has been verified through additional experiments. 
% Looking ahead, 
While we currently use all codewords for linear combination, we aim to explore sparse representations in the future work.
Besides, we plan to extend our research scope by applying ACCEPT to a wider range of tasks with a more diverse set of LLMs.

\section*{Limitations}
% Limitation is not counted in the 8 page limit, but it's mandatory
% 兩點: 多了 sub_dim 這個 hyperparam, large model 的 limitation 
While our extensive experiments across 17 datasets highlight the effectiveness of ACCEPT, it's important to acknowledge some additional considerations. Our method introduces some extra hyperparameters, such as determining the optimal sub-dimension $t$, which requires some extra %more 
computational efforts. Moreover, ACCEPT involves managing two distinct learning rates for \textit{SCPP} and \textit{SCAP}. Additionally, due to the significant resource requirements of the models with tens of billions of parameters, our experiments were conducted on a limited number of datasets. Future work will aim to explore ACCEPT on a broader range of datasets and larger models to further validate its efficacy.

% \section*{Acknowledgements}
% %%%%%%%%%%%%%%%%%%%%%%%
% This work was supported in part under grants {\color{black}NSTC 112-2634-F-002-005, NSTC112-2634-F-006-002, NSTC 112-2221-E-002 -132 -MY3, and NTU under grants 113L900902}.

\bibliography{custom}

\clearpage
\appendix

{\noindent\large\textbf{Appendix}}\label{sec:appendix}

%\begin{table*}[h!]
\begin{table*}[t]
  \centering
  % \tabcolsep=0.08cm
  % \small
  {\resizebox{\linewidth}{!}{
  \begin{tabular}{l|crrrccc}
    \toprule
    Dataset Name & Benchmark & \#Train & \#Valid & \#Test & Task Type & Metric \\
    \hline 
    MNLI & GLUE & 392,702 & 9,832 & 9,815 & Natural Language Inference (NLI) & accuracy \\
    QQP & GLUE & 362,846 & 1,000 & 40,431 & Paraphrase Detection & \underline{accuracy}/F1 \\
    QNLI & GLUE & 103,743 & 1,000 & 5,463 & NLI & accuracy \\
    SST-2 & GLUE & 66,349 & 1,000 & 872 & Sentiment Analysis & accuracy \\
    STS-B & GLUE & 5,749 & 750 & 750 & Sentence Similarity & \underline{Pearson}/Spearman corr. \\
    MRPC & GLUE & 3,668 & 204 & 204 & Paraphrase Detection & \underline{accuracy}/F1\\
    RTE & GLUE & 2,490 & 138 & 139 & NLI & accuracy \\
    CoLA & GLUE & 8,551 & 521 & 522 & Acceptability & Matthews corr. \\
    MultiRC & SuperGLUE & 27,243 & 2,424 & 2,424 & Question Answering (QA) & \underline{F1}/EM\\
    BoolQ & SuperGLUE & 9,427 & 1,635 & 1,635 & Boolean QA & accuracy \\
    WiC & SuperGLUE & 5,428 & 319 & 319 & Word Sense Disambiguation & accuracy \\
    WSC & SuperGLUE & 554 & 52 & 52 & Commonsense Reasoning & accuracy \\
    CB & SuperGLUE & 250 & 28 & 28 & NLI & accuracy \\
    ReCoRD & SuperGLUE & 137,484 & 1,370 & 15,176 & Commonsense Reasoning & \underline{F1}/EM \\
    NaturalQuestions & MRQA 2019 & 103,071 & 1,000 & 12,836 & Extractive QA & \underline{F1}/EM\\
    HotpotQA & MRQA 2019 & 71,928 & 1,000 & 5,901 & Extractive QA & \underline{F1}/EM\\
    SearchQA & MRQA 2019 & 116,384 & 1,000 & 16,980 & Extractive QA & \underline{F1}/EM\\
    NewsQA & MRQA 2019 & 73,160 & 1,000 & 4,212 & Extractive QA & \underline{F1}/EM \\
    SQuAD & MRQA 2019 & 86,599 & 1,000 & 10,570 & Extractive QA & \underline{F1}/EM\\
    \bottomrule
    \end{tabular}}}
    \caption{Detailed information of all datasets used in our experiments. For datasets that originally use two metrics, we designate the underlined metric as our primary evaluation measure following prior works~\citep{attempt:22,dept:24}.}  \label{tab:dataset}
\end{table*}

\begin{table*}[ht!]
  \centering
  % \tabcolsep=0.03cm
  % \small
  {\resizebox{1\linewidth}{!}{
  \begin{tabular}{l|cccccccgc}
    \toprule
    \multicolumn{10}{c}{\textbf{Soft-weighted Codebook Prepended Prompt (\textit{SCPP})}} \\
    \bottomrule
    \textbf{(t, r)} & 
    % (8, 7) &
    (16, 12)  &
    \cellcolor{lightlightgray}(32, 20) & (64, 30) & \cellcolor{lightlightgray}(96, 36) &
    (128, 40) & (192, 45) & (256, 48) &
    (384, 51) & (768, 55) \\
    \hline 
    \textbf{\#Para.} & 
    % 76416 &
    74496 & 
    \cellcolor{lightlightgray}74880 & 75360 & \cellcolor{lightlightgray}75648 &
    75840 & 76080 & 76224 & 76008 & 76260 \\
    \hline
    \textbf{Acc.} & 
    % 77.70 &
    77.70 &
    \cellcolor{lightlightgray}\textbf{82.73} & 79.14 & \cellcolor{lightlightgray}\textbf{82.73} &
    79.14 & 81.29 & 77.70 & \textbf{82.73} & 81.29 \\
    \toprule    
    \multicolumn{10}{c}{\textbf{Soft-weighted Codebook Added Prompt (\textit{SCAP})}} \\
    \bottomrule
    \textbf{(t, r)} & 
    % (8, 1) &
    (16, 2) & (32, 4) & (64, 8) & (96, 10) &
    \cellcolor{lightlightgray}(128, 13) &(192, 17) & \cellcolor{lightlightgray}(256, 20) & (384, 24) & (768, 30) \\
    \hline 
    \textbf{\#Para.} &
    % 71424 &
    72192 & 73728 & 76800 & 74240 &
    \cellcolor{lightlightgray}76032 & 76544 & \cellcolor{lightlightgray}76800 & 76800 & 76800 \\
    \hline
    \textbf{Acc.} &
    % 77.70 &
    78.42 & 77.70 & 76.98 & 76.98 &
     \cellcolor{lightlightgray}79.86 & 78.42 & \cellcolor{lightlightgray}81.29 & \textbf{82.73} & 78.42 \\
    \hline
    \end{tabular}}}
    \caption{Performance on RTE dataset with dividing the \textit{SCPP} and \textit{SCAP} into different granularities. Note that for T5-base, $t = 768$ means the prompt is NOT divided.}
  \label{tab:abla_pq_whole}
\end{table*}

\begin{table*}[t!]
  \centering
  \tabcolsep=0.08cm
  \small
  \begin{tabular}{l|ccccccccg|cccccg}
    \hline
    \multirow{2}{*}{\textbf{Init. Method}} & \multicolumn{9}{c}{\textbf{GLUE}} & \multicolumn{6}{c}{\textbf{SuperGLUE}} \\
    \cline{2-16}
    & MNLI & QQP & QNLI & SST-2 & STS-B & MRPC & RTE & CoLA & Avg. & Multi & Bool & WiC & WSC & CB & Avg. \\
    \hline 
    Random & 85.7 & 90.2 & 93.0 & 94.3 & \textbf{91.0} & \textbf{93.1} & 84.2 & 65.2 & 87.1 & 74.5 & 81.0 & \textbf{70.5} & \textbf{67.3} & 92.9 & 77.2 \\
    Intermediate task & \textbf{85.9} & 90.2 & \textbf{93.3} & 94.2 & \textbf{91.0} & 92.7 & \textbf{86.3} & 66.4 & \textbf{87.5} & 73.5 & \textbf{82.3} & 68.7 & \textbf{67.3} & \textbf{96.4} & \textbf{77.6} \\
    Target task & \textbf{85.9} & \textbf{90.4} & 93.1 & \textbf{94.5} & \textbf{91.0} & 91.7 & 81.3 & \textbf{68.8} & 87.1 & \textbf{74.9} & 81.8 & 69.0 & \textbf{67.3} & 92.9 & 77.2 \\   
    \hline
    \end{tabular}
    \caption{Performance on GLUE and SuperGLUE with different prompt initialization.}
  \label{tab:init_glue}
\end{table*}

\begin{table*}[ht!]
  \centering
  \tabcolsep=0.08cm
  \small
  % {\resizebox{0.9\linewidth}{!}{
  \begin{tabular}{lc|ccccccccg|cccccg}
    \toprule
    \multirow{2}{*}{\textit{SCPP}} & 
    \multirow{2}{*}{\textit{SCAP}} & \multicolumn{9}{c}{\textbf{GLUE}} & \multicolumn{6}{c}{\textbf{SuperGLUE}} \\
    \cline{3-16}
    && MNLI & QQP & QNLI & SST-2 & STS-B & MRPC & RTE & CoLA & Avg. & Multi & Bool & WiC & WSC & CB & Avg. \\
    \hline 
    \xmark & \xmark & 85.0 & \textbf{90.4} & \textbf{93.2} & 94.2 & 90.8 & 90.7 & 79.1 & 63.8 & 85.9 & 74.3 & 79.3 & 68.7 & \textbf{67.3} & 92.9 & 76.5 \\
    \checkmark & \xmark & 85.9 & 90.3 & \textbf{93.2} & 94.3 & 91.0 & 91.7 & 82.7 & 67.5 & 87.1 & 74.3 & 80.9 & 70.2 & \textbf{67.3} & \textbf{96.4} & 77.8 \\
    \xmark & \checkmark & \textbf{86.0} & \textbf{90.4} & \textbf{93.2} & 94.3 & \textbf{91.1} & 90.7 & 82.7 & 66.8 & 87.0 & \textbf{75.4} & 81.2 & 67.4 & \textbf{67.3} & \textbf{96.4} & 77.5 \\ 
    \checkmark & \checkmark & 85.9 & \textbf{90.4} & 93.1 & \textbf{94.5} & 91.0 & \textbf{93.1} & \textbf{86.3} & \textbf{68.8} & \textbf{87.9} & 74.9 & \textbf{82.3} & \textbf{70.5} & \textbf{67.3} & \textbf{96.4} & \textbf{78.3} \\
    \bottomrule
    \end{tabular}%}}
    \caption{Ablation study of \textit{SCPP} and \textit{SCAP} on GLUE and SuperGLUE benckmarks. We provide the performance of each dataset.}
  \label{tab:abla_glues}
\end{table*}

\section{Experimental Setting}\label{appendix:hyper}
We use PyTorch\footnote{\url{https://pytorch.org/}}, huggingface transformers\footnote{\url{https://github.com/huggingface/transformers}} and huggingface PEFT\footnote{\url{https://github.com/huggingface/peft}} to implement our work. GLUE\footnote{\url{https://huggingface.co/datasets/glue}}, SuperGLUE\footnote{\url{https://huggingface.co/datasets/super_glue}} and MRQA 2019 
Shared Task\footnote{\url{https://huggingface.co/lucadiliello}} are downloaded from huggingface dataset. We use the original T5 checkpoint rather than the LM-adapted 1.1 version~\citep{pt:21}. We modified codes based on DePT's repository\footnote{\url{https://github.com/ZhengxiangShi/DePT}}. We mainly cite the experiment results from ~\citet{tpt:23} and ~\citet{dept:24}. We typically use $m=60$ for the length of \textit{SCPP}, and set the maximum sequence length $l$ to $256$, which also corresponds to the length of \textit{SCAP} (except using $348$ for MultiRC following~\citet{dept:24}). We partition \textit{SCPP} and \textit{SCAP} into $K=24$ and $K=2$ subsections, respectively. The associated $r$ is calculated by the equation $rd + rmK \leq md$ for each model with dimension $d$. 
% We primarily use a grid search to determine the learning rates (lr) for both the codebook and weights in \textit{SCPP} and \textit{SCAP}. For \textit{SCPP}, the lr searched are \{3e-1, 4e-1, 5e-1\}, while for \textit{SCAP}, we searched \{1e-4, 5e-4, 1e-3, 5e-3\}. 
% Additionally, we observe that a larger lr is more suitable for \textit{SCAP} on the MRQA 2019 Shared Task. Therefore, we extend our search to include higher values $\{1, 5, 10\}$ for \textit{SCAP}. Note that for the experiments that train \textit{SCPP} or \textit{SCAP} alone, the backbone follows DePT (length of prompt is $60$ and rank of LoRA matrices equals to $30$).
% We train $30k$ steps for small datasets with less than $10k$ samples, and $300k$ steps for large datasets more than $10k$ samples, following~\citet{spot:21}. 
% We perform evaluations every 1,000 steps and save the best checkpoint based on performance on the evaluation dataset. The results on the test dataset are then reported using these best checkpoints. We choose a batch size of $16$ for T5-small, T5-base and T5-large models, $2$ for T5-3B and $1$ for Flan-T5-11B due to the GPU memory limitation. The warmup step and weight decay are $1,800$ and $0.01$, respectively. 
% Experiments are conducted on a single Nvidia 3090 GPU with 24 GB memory or multiple Nvidia V100 GPUs with 32 GB memory. 
{\color{black} As for the experiments using the Llama-2-7B model, we modified codes based on Petals' repository\footnote{\url{https://github.com/bigscience-workshop/petals}}. We use a learning rate of 3e-3 for SCPP and 5e-5 for SCAP. The weight decay is 
% set to 
1e-2 and 1e-3, respectively, with a batch size of 32.} %Experiments are conducted on a single Nvidia 3090 GPU with 24 GB memory or multiple Nvidia V100 GPUs with 32 GB memory.}

\section{Task and Dataset Details}
We list the detailed information, including numbers of training, evaluation and testing samples, task types and evaluation metrics of each dataset which has been used in our experiments in Tab.~\ref{tab:dataset}. We utilize a diverse range of datasets covering various NLU tasks, including Natural Language Inference (NLI), Paraphrase Detection, and Sentiment Analysis. Additionally, we explore different types of Question Answering (QA) tasks, such as extractive and boolean QA. The effectiveness and generalizability of ACCEPT are demonstrated across these tasks in Tab.~\ref{tab:glue&superglue}
 and Tab.~\ref{tab:mrqa}.

\section{More Details of Experiments}
\label{appendix:method}
In this section, we present more comprehensive experiments.

\begin{figure}[t!]
\includegraphics[width=1\columnwidth]{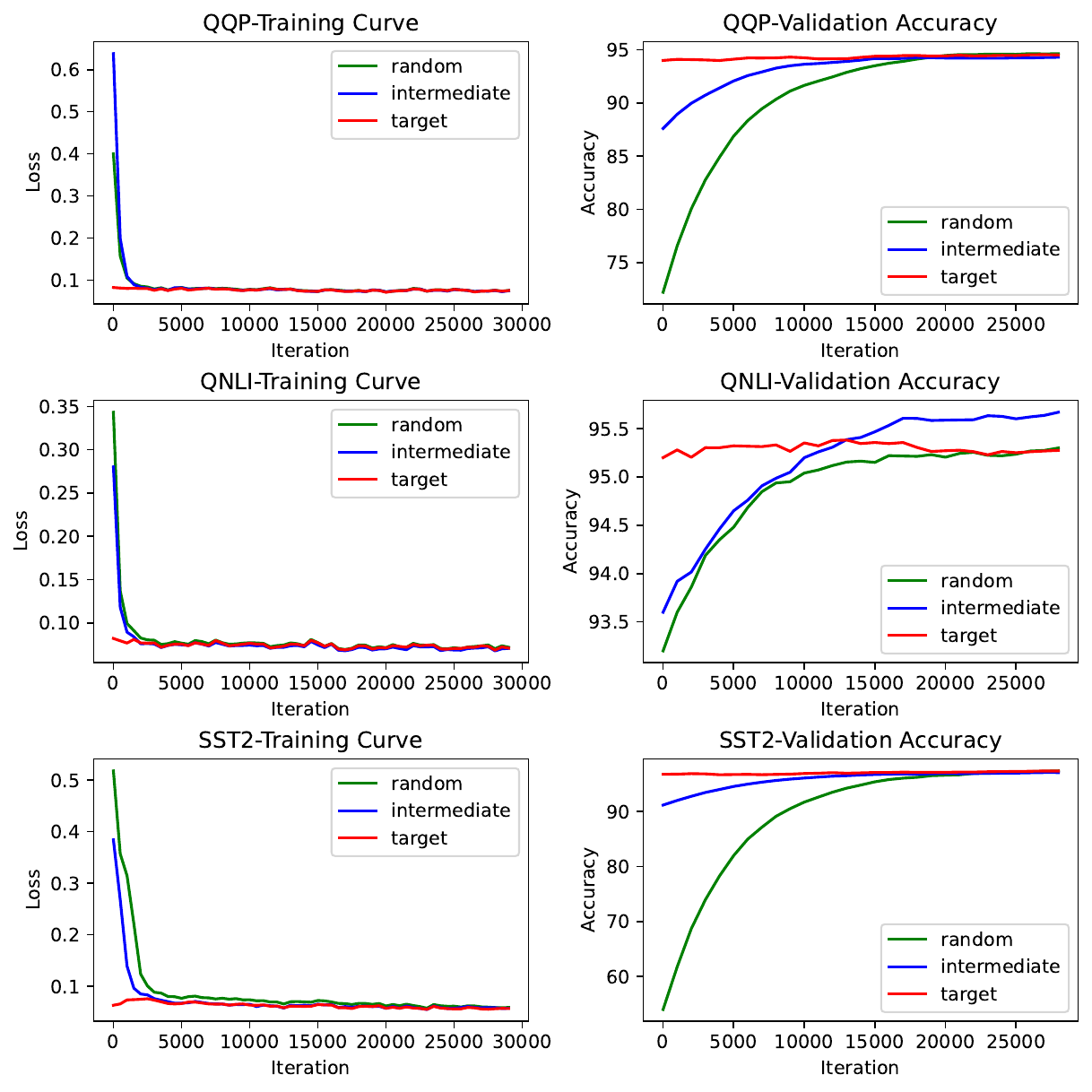}
  \caption{Training curve (left) and validation accuracy curve (right) comparison between different prompt initialization strategies across QQP, QNLI and SST-2.}
  \label{fig:training_curve}
\end{figure}

\subsection{Details of Prompt Initialization}
\label{appendix:init}
Tab.~\ref{tab:init_glue} and Tab.~\ref{tab:init_mrqa} present the results for each dataset using three initialization strategies. The majority of performances improved with either intermediate task initialization or target task initialization, demonstrating the effectiveness of pre-learning knowledge before transferring it to the target tasks, aligning with SPoT~\cite{spot:21}. 

{\color{black} In addition, we present a comparison of the training curves and validation curves using different methods of prompt initialization across QQP, QNLI and SST-2 datasets, as shown in Figure~\ref{fig:training_curve}. It can be observed that initializing with an intermediate task or target task helps the target task transfer more easily, resulting in faster convergence and better performance.}

\subsection{Details of Different Granularity of Subdivision.}
\label{appendix:abla_gran}
We have shown the performance of different sub-dimension ($t$) and codebook size ($r$) pairs in Tab.~\ref{tab:abla_gran}. We present more results in Tab.~\ref{tab:abla_pq_whole} by selecting total $8$ factors of the model dimension ($d = 768$ for T5-base) and conduct the experiments for each setting on RTE dataset. Tab.~\ref{tab:abla_pq_whole} shows that with an appropriate division, multiple configurations surpass the performance of not dividing prompts ($t = 768$), which demonstrate the effectiveness of PQ. We select $t= 32$, $K=24$, $r=20$) for \textit{SCPP} and $t=384$, $K=2$, $r=24$) for \textit{SCAP} as the final decision considering both the performance and parameter efficiency, applying to all datasets.

\subsection{Details of Ablation on \textit{SCPP} and \textit{SCAP}.}
\label{appendix:abla_pq}
In the main paper, we provide the average performances on GLUE/SuperGLUE benchmarks and MRQA 2019 Shared Task in Tab.~\ref{tab:pre_add_abla}. Here we provide the performance on each dataset in Tab.~\ref{tab:abla_glues} and Tab.~\ref{tab:abla_mrqa}. The results of most datasets show improvements when using either \textit{SCPP} or \textit{SCAP} individually, and are the best performances when both are applied simultaneously, further validating the effectiveness of ACCEPT.

\begin{table}[t]
  \centering
  \tabcolsep=0.08cm
  \small
  \begin{tabular}{l|ccccg}
    \hline
    \multirow{2}{*}{\textbf{Init. Method}} & \multicolumn{5}{c}{\textbf{MRQA}} \\
    \cline{2-6}
     & NQ & HP & SQA & News & Avg. \\
    \hline 
    % Random & 72.97 & 76.59 & 78.27 & 64.63 & 73.115 \\
    Random & 73.47 & 76.74 & 78.59 & \textbf{64.63} & 73.36 \\
    Intermediate task & 72.71 & 76.98 & 78.47 & 64.44 & 73.15  \\
    Target task & \textbf{73.61} & \textbf{77.10} & \textbf{78.91} & 64.62 & \textbf{73.55} \\
    \hline
    \end{tabular}
    \caption{Performance on MRQA 2019 Shared Task with different prompt initialization.}
  \label{tab:init_mrqa}
\end{table}

%%%%%%%%%%%%%%%%%%%%%%%
%\begin{table}[t!]
\begin{table}[t]
  \centering
  % \tabcolsep=0.04cm
  % \small
  {\resizebox{1\linewidth}{!}{
  \begin{tabular}{ccccccg}
    \hline
    \multirow{2}{*}{\textit{SCPP}} & 
    \multirow{2}{*}{\textit{SCAP}} & \multicolumn{5}{c}{\textbf{MRQA}} \\
    \cline{3-7} 
    && NQ & HP & SQA & News & Avg. \\
    \hline
     \xmark & \xmark & $73.2_{0.1}$ & $76.8_{0.3}$ & $77.6_{0.2}$ & $64.4_{0.1}$ & 73.0 \\
    % V & X & 73.8 & 76.7 & 79.1 & 64.5 & 73.5 \\
    % X & V & 73.3 & 76.5 & 78.4 & 64.4 & 73.2 \\
    % V & V & 73.6 & 77.0 & 78.9 & 64.6 & 73.5 \\
    \checkmark & \xmark & \color{black}73.8$_{0.05}$ & {\color{black}76.9$_{0.01}$} & \color{black}78.8$_{0.2}$ & {\color{black}64.7$_{0.1}$} & \color{black}73.5 \\
    \xmark & \checkmark & {\color{black}73.4$_{0.2}$} & \color{black}76.8$_{0.5}$ & \color{black}78.5$_{0.2}$ & \color{black}64.3$_{0.1}$ & \color{black}73.3 \\
    \checkmark & \checkmark & \color{black}73.6$_{0.05}$ & \color{black}77.1$_{0.1}$ & \color{black}78.9$_{0.01}$ & \color{black}64.6$_{0.06}$ & \color{black}\textbf{73.6} \\
    \hline
    \end{tabular}}}
    \caption{Ablation study of \textit{SCPP} and \textit{SCAP} on MRQA 2019 Shared Task. We report the average F1 and standard deviation of three experiments with different seeds.}
  \label{tab:abla_mrqa} 
\end{table}

\end{document}